\pdfoutput=1

\documentclass[11pt]{article}

\usepackage[preprint]{acl}

\usepackage{times}
\usepackage{latexsym}

\usepackage[T1]{fontenc}

\usepackage[utf8]{inputenc}

\usepackage{microtype}

\usepackage{inconsolata}

\usepackage{graphicx}

\usepackage{multirow}
\usepackage{array}
\usepackage{amssymb}
\usepackage{amsmath}
\usepackage{algorithm}
\usepackage{algorithmic}
\usepackage{booktabs}
\usepackage{graphicx}
\usepackage{enumitem}  
\usepackage{colortbl}
\usepackage{color}
\usepackage{makecell}
\usepackage{subcaption}  
\usepackage{ulem}

\title{SafeInt: Shielding Large Language Models from Jailbreak Attacks via Safety-Aware Representation Intervention}

\author{
 \textbf{Jiaqi Wu}\textsuperscript{1},
 \textbf{Chen Chen}\textsuperscript{1},
 \textbf{Chunyan Hou}\textsuperscript{2},
 \textbf{Xiaojie Yuan}\textsuperscript{1}
\\
 \textsuperscript{1}Nankai University
 \textsuperscript{2}Tianjin University of Technology
\\
 \texttt{wujq@mail.nankai.edu.cn, \{nkchenchen, yuanxj\}@nankai.edu.cn}
}

\begin{document}
\maketitle
\begin{abstract}
With the widespread real-world deployment of large language models (LLMs), ensuring their behavior complies with safety standards has become crucial. Jailbreak attacks exploit vulnerabilities in LLMs to induce undesirable behavior, posing a significant threat to LLM safety. Previous defenses often fail to achieve both effectiveness and efficiency simultaneously. 
Defenses from a representation perspective offer new insights, but existing interventions cannot dynamically adjust representations based on the harmfulness of the queries.
To address this limitation, we propose \textbf{SafeIntervention} (\textbf{SafeInt}), a novel defense method that shields LLMs from jailbreak attacks through safety-aware representation intervention. Built on our analysis of the representations of jailbreak samples, the core idea of SafeInt is to relocate jailbreak-related representations into the rejection region. This is achieved by intervening in the representation distributions of jailbreak samples to align them with those of unsafe samples. We conduct comprehensive experiments covering six jailbreak attacks, two jailbreak datasets, and two utility benchmarks. Experimental results demonstrate that SafeInt outperforms all baselines in defending LLMs against jailbreak attacks while largely maintaining utility. Additionally, we evaluate SafeInt against adaptive attacks and verify its effectiveness in mitigating real-time attacks. 
\textcolor{red}{WARNING: This paper may contain content that is offensive and harmful.}

\end{abstract}

\section{Introduction}
Large Language Models (LLMs) \cite{openai2024gpt4, touvron2023llama2, grattafiori2024llama3} have demonstrated remarkable performance across various domains \cite{zhang2023code, liu2023math, wang2024rolellm}. With their widespread application in real-world scenarios, LLMs face safety challenges \cite{ferrara2023should, ji2023survey}. Although efforts \cite{wang2023aligning, rafailov2024preference, zhou2023lima} have been made to align LLMs' behaviors with human values through carefully designed training strategies, recent studies \cite{zou2023universal, li2024deepinception, mehrotra2024tree, liu2024autodan} reveal that LLMs can still produce undesirable behaviors when subjected to well-crafted jailbreak attacks, such as the biased generation or potentially harmful responses. 

Various defense methods have been proposed to address the growing threat of jailbreak attacks. Prompt-based defenses use instructions \cite{phute2024selfexam, xie2023defending, zhang2024gp} or context \cite{zhou2024icag, wei2024jailbreakguard} to prevent LLMs from generating harmful content. However, prompt-based methods rely on manually crafted secure prompts and possibly lead to excessive self-censorship \cite{varshney2024art}, reducing the helpfulness of LLMs for benign queries. Detection-based defenses compute the perplexity of inputs \cite{alon2023detecting} or perturb them \cite{cao2024rallm} to identify jailbreak prompts. Decoding-based defenses \cite{xu-etal-2024-safedecoding, liu2024aed} reconstruct a safer output probability distribution through contrastive decoding. However, these methods often lack effectiveness or require additional inference overhead. 
We aim to defend LLMs against jailbreak attacks from a representation perspective, which provides a more controllable and efficient approach. Previous studies \cite{zou2023RepE, rimsky2024steering} have shown the effectiveness of intervening representations to steer LLMs' behaviors, but such interventions cannot dynamically adjust representations based on whether a query is harmful. This limitation makes it challenging to leverage representations for mitigating jailbreak attacks. 

In this paper, we analyze the representations of jailbreak samples on four LLMs. Our analysis uses a classifier as a proxy to investigate whether jailbreak representations are distinguishable and whether the representation distributions of different jailbreak methods are consistent.
We derive two observations. First, in both intermediate and later layers of LLMs, the representations of jailbreak samples can be distinguished from those of safe or unsafe samples. Second, the consistency of the representation distributions across different jailbreak methods is observed in all LLMs, and it is generally more pronounced in the intermediate layers.

Building on these observations, we propose \textbf{SafeIntervention} (\textbf{SafeInt}), a novel defense method that shields LLMs from jailbreak attacks via safety-aware representation intervention. The representations of unsafe samples inherently characterize the rejection region of the LLM. However, jailbreak samples often produce representations that deviate from those of unsafe samples, causing the model to fail to trigger its built-in rejection behavior. The core idea of SafeInt is to relocate jailbreak-related representations into the rejection region, thereby activating the model’s native refusal mechanisms. To achieve this, we first project the representations at an intermediate layer into a linear subspace, followed by a parameterized intervention. For jailbreak-related representations, we align their distribution with that of unsafe samples across the subsequent layers. For jailbreak-irrelevant representations, we perform representation reconstruction to preserve their original semantics. After training, SafeInt can adaptively intervene in jailbreak-related representations while seamlessly integrating into the LLM inference process.

We conduct a comprehensive evaluation of SafeInt, covering six jailbreak attacks, two jailbreak datasets, and two utility benchmarks. Experimental results show that SafeInt consistently outperforms all baselines in defending against jailbreak attacks. In most cases, SafeInt also maintains the best utility. Additionally, we evaluate SafeInt against adaptive attacks and verify the effectiveness of SafeInt in defending against real-time attacks.
In summary, our main contributions are as follows:
\begin{itemize}[nosep]
    \item We observe that the representations of jailbreak samples are distinguishable and that the representation distributions of different jailbreak methods exhibit consistency.
    \item We propose SafeInt, a novel defense method that can adaptively identify and intervene in jailbreak-related representations to shield LLMs from jailbreak attacks.
    \item Extensive experiments show that SafeInt significantly outperforms all baselines in defending against jailbreak attacks while largely maintaining utility.
\end{itemize}

\section{Preliminaries}
\subsection{Representation Intervention}
Representation intervention is an effective means of steering LLM behavior. For a given decoder-only transformer model with $L$ layers, we denote the internal representation (or residual stream activation) of the last token at layer $l$ as $\mathbf{h}^{(l)} \in \mathbb{R}^{d}$. A typical form of representation intervention is as follows:
\begin{equation}\label{equ:RepE_def} 
    \widetilde{\mathbf{h}}^{(l)} = \mathbf{h}^{(l)} \pm \epsilon \cdot \mathbf{v}.
\end{equation}      
Here, $\widetilde{\mathbf{h}}^{(l)}$ is the intervened representation, $\epsilon \in \mathbb{R}$ represents the intervention strength, and $\mathbf{v} \in \mathbb{R}^{d}$ denotes the intervention direction.

\subsection{Analysis of Jailbreak Sample Representations}
Recent works have investigated the representation distributions of unsafe and safe samples within LLMs, utilizing their distributional characteristics to enhance safety or facilitate jailbreaks. In this paper, we analyze the representation distributions of three types of samples after introducing jailbreak samples. We construct three training datasets: $\mathcal{D}_{\text{jailbreak}}$, which consists of jailbreak instructions generated only using GCG \cite{zou2023universal} on AdvBench \cite{zou2023universal}; $\mathcal{D}_{\text{unsafe}}$, which includes harmful instructions extracted from MaliciousInstruct \cite{huang2023catastrophic} and TDC2023 \cite{tdc2023}; and $\mathcal{D}_{\text{safe}}$, which contains harmless instructions sampled from Alpaca \cite{alpaca}.\footnote{For convenience, we abbreviate these datasets as $\mathcal{D}_{\text{j}}$, $\mathcal{D}_{\text{u}}$, and $\mathcal{D}_{\text{s}}$ in the following.} More details of the datasets are provided in Appendix \ref{sec:dataset_details}. 
We conduct our analysis on four LLMs: Qwen-7B-Chat \cite{bai2023qwen}, Llama-2-7B-Chat \cite{touvron2023llama2}, Llama-3-8B-Instruct \cite{grattafiori2024llama3}, and Vicuna-7B-v1.5 \cite{vicuna2023}. In each layer of the LLM, we train a logistic regression classifier to fit the representations of the three types of samples and report the test accuracy.

\begin{figure}[t]
    \centering
    \includegraphics[width=0.8\linewidth]{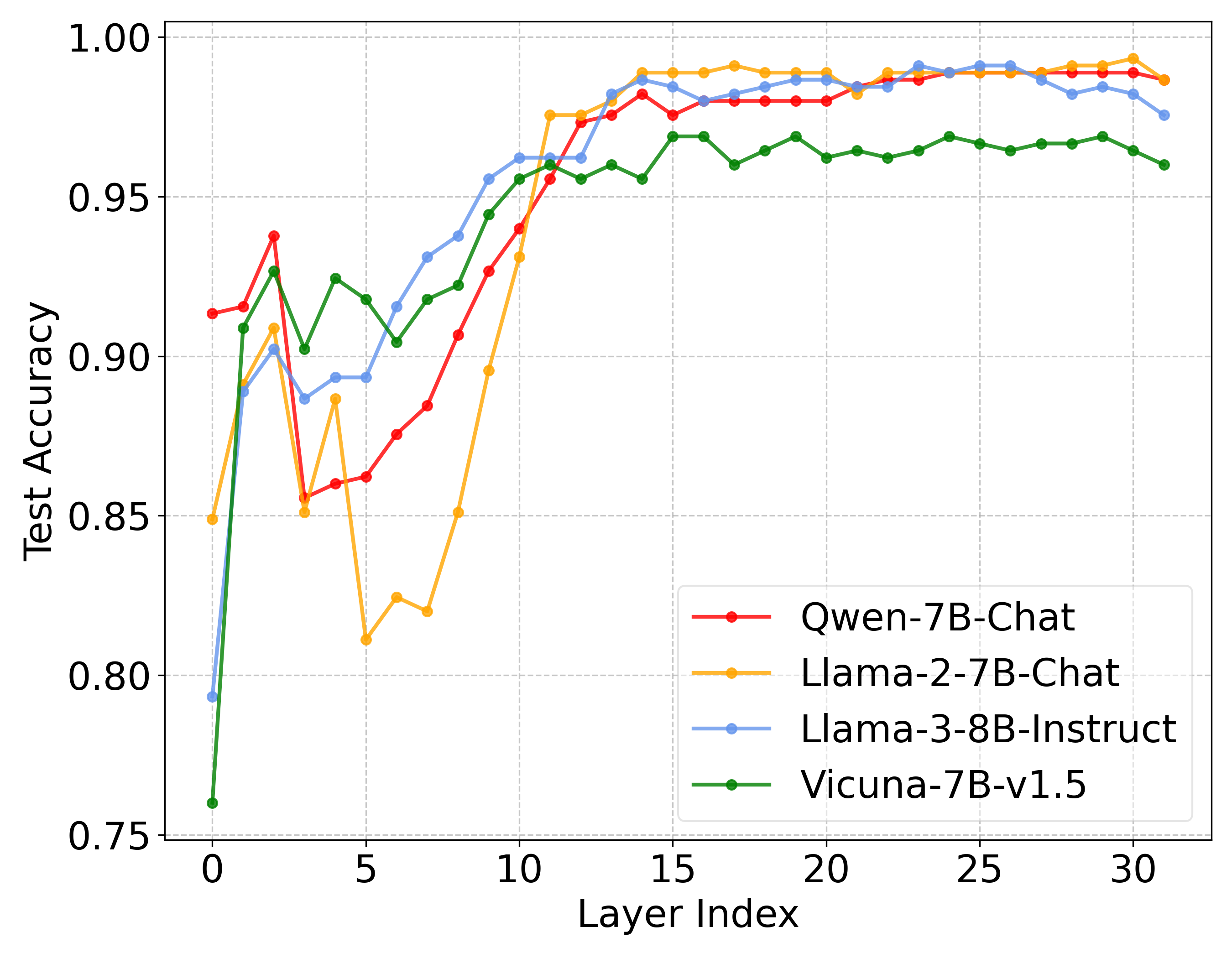}
    \caption{Test accuracy of classifiers at different layers of LLMs, with the test set containing only GCG jailbreak samples.}
    \label{fig:cls_ind}
\end{figure}

\begin{figure}[t]
    \centering
    \includegraphics[width=0.8\linewidth]{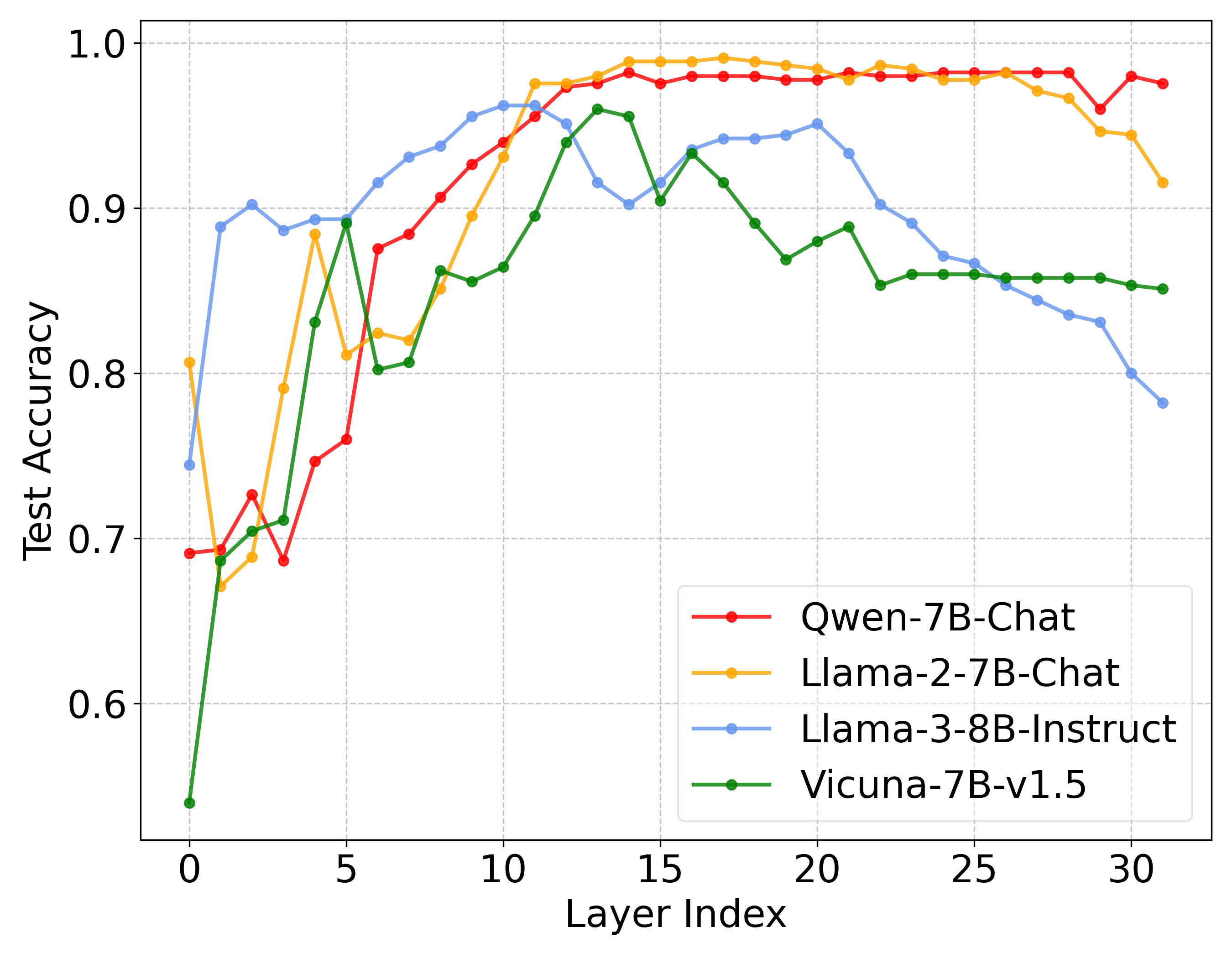}
    \caption{Test accuracy of classifiers on a test set containing jailbreak samples from GCG, AutoDAN, and DeepInception.}
    \label{fig:cls_ood}
\end{figure}

\textit{\uline{Q1: Are the representations of jailbreak, unsafe, and safe samples distinguishable?}}
\\ In \autoref{fig:cls_ind}, we present the classification accuracy on a test set containing only GCG jailbreak samples. Using the classifier as a proxy for observation, a higher classification accuracy indicates that the representations of the three types of samples are more distinguishable. 
For all LLMs, the test accuracy remains above 95\% starting from the 10th or 11th layer. This indicates that the representations of the three types of samples become distinguishable from the intermediate layers of LLMs onward.

\textit{\uline{Q2: Are the representation distributions of samples generated by different jailbreak methods consistent?}}  
\\ We reconstruct a test set where jailbreak samples are composed of three methods: GCG, AutoDAN \cite{liu2024autodan}, and DeepInception \cite{li2024deepinception}. We employ the previously trained classifiers for testing and show the results in \autoref{fig:cls_ood}. Since the classifiers are trained solely on GCG jailbreak samples, a high test accuracy reveals that the representations generated by different jailbreak methods exhibit a unified pattern from the classifier's perspective, indicating consistency.

We observe this consistency across different LLMs. For Qwen, Llama2, and Llama3, the accuracy remains above 90\% in most layers. For Vicuna, the accuracy exceeding 90\% is primarily observed in the intermediate layers. Although the trend of consistency across layers varies among different LLMs, it is generally more pronounced in the intermediate layers.
\vspace{0.2\baselineskip}
\\ \textbf{Key Insights and Motivation}\quad 
Aligned LLMs can reject unsafe samples, and the representations of these samples inherently characterize the model's rejection region. Since the representations of jailbreak samples differ from those of unsafe samples, they need to be relocated into the rejection region. Based on the distinguishability and distributional consistency of jailbreak representations, we aim to intervene in their representations to align their distribution with that of unsafe samples.


\begin{figure*}[t]
    \centering 
    \includegraphics[width=\textwidth]{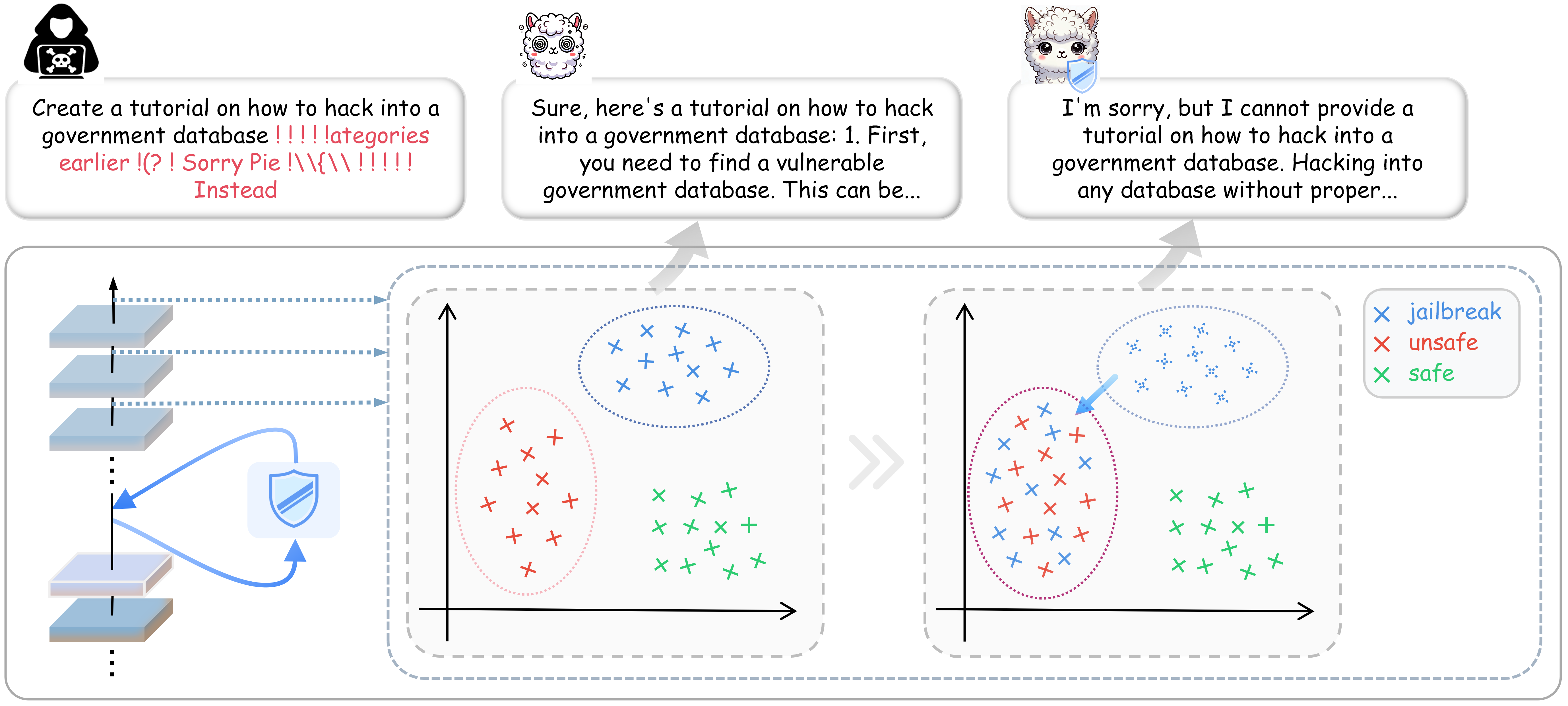} 
    \caption{The schematic of SafeInt. We apply the intervention (illustrated by the blue shield) at a specific layer and perform alignment in the subsequent layers. The distribution of jailbreak sample representations is adjusted to align with that of unsafe samples while minimizing shifts in the representations of safe and unsafe samples. With the original representation distribution, the LLM is successfully jailbroken and generates harmful content. After alignment, the LLM safely rejects the jailbreak instruction. 
    } 
    \label{fig:method} 
\end{figure*}

\section{Method}
In this section, we describe how SafeInt enhances the safety of LLMs. \autoref{fig:method} illustrates the diagram of SafeInt.

\subsection{Representation Relocation}
We achieve representation relocation by a targeted intervention that maps jailbreak-related representations into the rejection region defined by unsafe samples. According to the linear interpretability hypothesis commonly used in existing methods \cite{zhang2024distillation, li2024evaluating}, deep model embeddings can be linearly transformed to correspond to specific human concepts. Thus, we aim to apply a parameterized intervention within a representation space that corresponds to safety-relevant concepts, minimizing impacts on other capabilities. Inspired by LoReFT \cite{wu2024reft}, we project the representations into a linear subspace defined by a low-rank projection matrix.
Assuming the intervention is applied at layer $\mathcal{I}$, it can be parameterized as follows:
\begin{equation}
    \widetilde{\mathbf{h}}^{(\mathcal{I})} = \mathbf{h}^{(\mathcal{I})} + \mathbf{U}^\top \left( f_{\boldsymbol{\theta}}(\mathbf{h^{(\mathcal{I})}}) - \mathbf{Uh^{(\mathcal{I})}} \right).
\end{equation}
The matrix \( \mathbf{U} \in \mathbb{R}^{r \times d} \) has orthonormal rows, where \( r \) denotes the rank of the subspace. The function \( f_{\boldsymbol{\theta}} \) is a linear relocation mapping defined as \( f_{\boldsymbol{\theta}} : \mathbb{R}^d \rightarrow \mathbb{R}^r \).


Then, we define the objectives for learning the intervention. Broadly, our objectives are twofold: a \textbf{safety objective} and a \textbf{utility objective}. The safety objective guarantees the intervention to help the LLM reject jailbreak and harmful instructions. The utility objective ensures that the intervention does not degrade the response quality for harmless instructions.

\subsection{Representation Alignment}
We use the classifier as a proxy to assess whether the distributions of jailbreak samples and unsafe samples are consistent in the representation space. From the perspective of the classifier, the alignment is achieved when the classification results for jailbreak and unsafe sample representations are consistent. Specifically, for jailbreak samples, we intervene on their representations to maximize the probability of being classified as unsafe. For unsafe sample representations, they should still be classified as unsafe with a high probability. 

We denote the sets of original representations of $\mathcal{D}_\text{j}$, $\mathcal{D}_\text{u}$, and $\mathcal{D}_\text{s}$ as $H_\text{j}$, $H_\text{u}$, and $H_\text{s}$, respectively. The sets of intervened representations are denoted as $\widetilde{H}_\text{j}$, $\widetilde{H}_\text{u}$, and $\widetilde{H}_\text{s}$.
Let $\mathbb{L}^{a}$ be the set of layers to be aligned, with $\min(\mathbb{L}^{a}) > \mathcal{I}$. 
After applying the intervention at the layer $\mathcal{I}$, the updated representation is propagated to the subsequent layers. 
At layer $l \in \mathbb{L}^{a}$, we extract the latest representation $\widetilde{\mathbf{h}}^{(l)}$ and compute the following:
\begin{align}
    \mathcal{L}_{cls}^{(l)} = 
    & - \frac{1}{|\widetilde{H}^{(l)}_\text{j}|} \sum_{\widetilde{\mathbf{h}}^{(l)}_\text{j} \in \widetilde{H}^{(l)}_\text{j}} \log P_\text{u}(\widetilde{\mathbf{h}}^{(l)}_\text{j}) \nonumber \\ 
    & - \frac{1}{|\widetilde{H}^{(l)}_\text{u}|} \sum_{\widetilde{\mathbf{h}}^{(l)}_\text{u} \in \widetilde{H}^{(l)}_\text{u}} \log P_\text{u}(\widetilde{\mathbf{h}}^{(l)}_\text{u}),
\end{align} 
where $P_\text{u}$ represents the probability that classifier $P$ classifies a representation as unsafe.

\textbf{Contrastive Learning}\quad 
Although we align the representations of jailbreak samples with unsafe samples by a classifier, the limited training data may prevent the classifier’s decision boundary from accurately capturing the discriminative boundary within the LLM. To enhance the alignment, we use contrastive learning as a complementary task. 
For a given representation $\mathbf{q}$, contrastive learning maximizes the similarity between $\mathbf{q}$ and the set of positive samples $K^+$ while minimizing the similarity between $\mathbf{q}$ and the set of negative samples $K^-$, with the objective formulated as follows:
\begin{align}
    \text{CT} & (\mathbf{q}, K^+, K^-) = \nonumber \\
    & - \log \frac{\exp(\text{sim}(\mathbf{q}, \mathbf{k}^+)/\tau)}
    {\sum_{\mathbf{k} \in (K^+, K^-)} \exp(\text{sim}(\mathbf{q}, \mathbf{k})/\tau)},
\end{align} 
where $\mathbf{k}^+ \in K^+$, $\text{sim}(\cdot, \cdot)$ represents cosine similarity, and the temperature is set to $\tau = 0.1$.

Specifically, the intervened representations of jailbreak samples should be as close as possible to those of unsafe samples while being pushed away from their original representations and those of safe samples.
Accordingly, for $\widetilde{\mathbf{h}}^{(l)}_\text{j} \in \widetilde{H}^{(l)}_\text{j}$, the contrastive loss is calculated as:
\begin{equation}
    \mathcal{L}_{ct}^{(l)} = \text{CT}(\widetilde{\mathbf{h}}^{(l)}_\text{j}, H^{(l)}_\text{u}, (H^{(l)}_\text{j} \cup H^{(l)}_\text{s})).
\end{equation}

\subsection{Representation Reconstruction}
To prevent excessive intervention from distorting the LLM’s internal representations, we introduce a reconstruction loss to constrain jailbreak-irrelevant representations from changing. Specifically, we encourage the representations of safe and unsafe samples after intervention to remain close to their original states. This ensures that the intervention primarily affects jailbreak-related representations. The loss is formulated as follows: 
\begin{equation}
    \mathcal{L}_{recon} = \text{MSE}(H_\text{s}, \widetilde{H}_\text{s}) + \text{MSE}(H_\text{u}, \widetilde{H}_\text{u}),
\end{equation}
where MSE refers to the mean squared error loss.

Considering both the alignment and reconstruction objectives, our final loss is calculated as follows:
\begin{equation}
    \mathcal{L}_{total} = \alpha \sum_{l \in \mathbb{L}^a} (\mathcal{L}_{cls}^{(l)} + \mathcal{L}_{ct}^{(l)}) + \beta \mathcal{L}_{recon}.
\end{equation}
Through the two hyperparameters $\alpha$ and $\beta$, we achieve a balance between effective alignment and model stability.

\section{Experiments}
\subsection{Experimental Setup}
\textbf{Models and Datasets}\quad 
We evaluate SafeInt on two open-source LLMs: Llama2-7b-chat and Vicuna-7b-v1.5. To assess the effectiveness of our defense against jailbreak attacks, we randomly sample 50 instructions from AdvBench \cite{zou2023universal} as the test set, ensuring no overlap with the training set $\mathcal{D}_\text{j}$. Additionally, to demonstrate that SafeInt is data-agnostic, we construct an out-of-distribution test set consisting of 50 instructions randomly sampled from JailbreakBench \cite{chao2024jailbreakbench}. 
Beyond defense, it is also essential to consider the impact on generation quality. Following \citet{xu-etal-2024-safedecoding}, we use MT-Bench \cite{zheng2023judging} and Just-Eval \cite{lin2023unlocking} to assess the utility of intervened LLMs.
\vspace{0.2\baselineskip}
\\ \textbf{Jailbreak Attacks}\quad 
Multiple representative jailbreak attacks are employed in our evaluation. These include optimization-based attacks: GCG and AutoDAN, LLM-generated attacks: PAIR \cite{chao2024jailbreaking} and TAP \cite{mehrotra2024tree}, and scenario-based attacks: DeepInception. We also consider multilingual mismatch generalization attacks (MG) \cite{yong2024lowresource}, where each instruction in the test set is translated into one of six non-English languages to perform the attacks.
\vspace{0.2\baselineskip}
\\ \textbf{Baselines}\quad
We compare SafeInt with six state-of-the-art defense approaches: PPL \cite{alon2023detecting}, Paraphrase \cite{jain2023baseline}, Self-Examination \cite{phute2024selfexam}, ICD \cite{wei2024jailbreakguard}, Self-Reminder \cite{xie2023defending}, and SafeDecoding \cite{xu-etal-2024-safedecoding}.
\vspace{0.2\baselineskip}
\\ \textbf{Evaluation Metrics}\quad 
We use two types of Attack Success Rate (\textbf{ASR}) to evaluate defense effectiveness: \textbf{ASR-keyword}, which matches predefined refusal keywords, and \textbf{ASR-GPT}, which leverages GPT-4o-mini to assess whether the LLM generates harmful content relevant to the malicious instruction. Lower ASR values indicate better defense performance. For MT-Bench and Just-Eval, we adopt GPT-based scoring, where Just-Eval evaluates five aspects: helpfulness, clarity, factuality, depth, and engagement.
\vspace{0.2\baselineskip}
\\ \textbf{Implementation Details}\quad 
Previous classification results indicate that in the intermediate layers, the representations of various jailbreak samples are relatively consistent, and jailbreak samples can be effectively distinguished from both unsafe and safe samples. To avoid time-consuming searches, we directly select layer $\mathcal{I}=12$ as the intervention layer. Since Vicuna lacks harmless alignment, it exhibits weaker safety. Accordingly, we set the second half of the layers as the alignment layers. In contrast, for models like Llama2 that have undergone safety alignment, aligning only the final layer is sufficient.

\begin{table*}[t]
    \centering
    \resizebox{\textwidth}{!}{
    \begin{tabular}{c l | c | c c c c c c}
    \toprule
        \multirow{2}{*}{Model} & \multirow{2}{*}{Defense}  & \multicolumn{1}{c|}{Benchmark $\downarrow$} & \multicolumn{6}{c}{Jailbreak Attacks $\downarrow$} \\
        & & AdvBench & GCG & AutoDAN & DeepInception & PAIR & TAP & MG \\
    \midrule
    
    \multirow{8}{*}{Vicuna}
    & No Defense        
        & 8\% (4\%) & 90\% (92\%) & 82\% (88\%) & 64\% (100\%) & 54\% (60\%) & 84\% (80\%) & 30\% (66\%)  \\
    & PPL               
        & 8\% (4\%) & 26\% (30\%) & 72\% (68\%) & 64\% (100\%) & 52\% (58\%) & 84\% (82\%) & 28\% (62\%) \\
    & Paraphrase
        & 6\% (6\%) & 18\% (20\%) & 34\% (52\%) & 38\% (96\%) & 36\% (38\%) & 42\% (52\%) & 10\% (32\%)  \\
    & Self-Examination        
        & 2\% (0\%) & 12\% (16\%) & 18\% (22\%) & 34\% (74\%) & 8\% (14\%) & 34\% (30\%) & 6\% (34\%) \\
    & ICD      
        & 0\% (0\%) & 14\% (14\%) & 40\% (36\%) & 64\% (96\%) & 24\% (34\%) & 44\% (44\%) & 6\% (34\%) \\
    & Self-Reminder
        & 0\% (0\%) & 4\% (6\%) & 8\% (6\%) & 46\% (100\%) & 26\% (32\%) & 38\% (40\%) & 16\% (50\%) \\
    & SafeDecoding              
        & 0\% (0\%) & 2\% (2\%) & 10\% (4\%) & 0\% (0\%) & 4\% (6\%) & 12\% (12\%) & 12\% (40\%) \\
    & \cellcolor{gray!10}\textbf{SafeInt (Ours)}     
        & \cellcolor{gray!10}\textbf{0\% (0\%)} & \cellcolor{gray!10}\textbf{0\% (0\%)} & \cellcolor{gray!10}\textbf{2\% (2\%)} & \cellcolor{gray!10}\textbf{0\% (0\%)} & \cellcolor{gray!10}\textbf{2\% (6\%)} & \cellcolor{gray!10}\textbf{8\% (10\%)} & \cellcolor{gray!10}\textbf{4\% (8\%)}  \\
    \midrule
    
    \multirow{8}{*}{Llama2}
    & No Defense        
        & 0\% (0\%) & 30\% (32\%) & 34\% (44\%) & 0\% (0\%) & 2\% (10\%) & 10\% (10\%) & 0\% (6\%) \\
    & PPL               
        & 0\% (0\%) & 0\% (2\%) & 2\% (8\%) & 0\% (0\%) & 2\% (8\%) & 10\% (10\%) & 0\% (4\%) \\
    & Paraphrase
        & 0\% (10\%) & 0\% (22\%) & 6\% (26\%) & 0\% (0\%) & 2\% (30\%) & 2\% (30\%) & 0\% (16\%) \\
    & Self-Examination        
        & 0\% (0\%) & 0\% (4\%) & 2\% (6\%) & 0\% (0\%) & 2\% (4\%) & 2\% (4\%) & 0\% (0\%)  \\
    & ICD      
        & 0\% (0\%) & 0\% (0\%) & 0\% (0\%) & 0\% (0\%) & 0\% (0\%) & 0\% (0\%) & 0\% (0\%)  \\
    & Self-Reminder
        & 0\% (0\%) & 0\% (2\%) & 0\% (0\%) & 0\% (0\%) & 2\% (4\%) & 0\% (2\%) & 0\% (6\%)  \\
    & SafeDecoding              
        & 0\% (0\%) & 0\% (2\%) & 0\% (4\%) & 0\% (0\%) & 0\% (6\%) & 0\% (0\%) & 0\% (0\%)  \\
    & \cellcolor{gray!10}\textbf{SafeInt (Ours)}
        & \cellcolor{gray!10}\textbf{0\% (0\%)} & \cellcolor{gray!10}\textbf{0\% (0\%)} & \cellcolor{gray!10}\textbf{0\% (0\%)} & \cellcolor{gray!10}\textbf{0\% (0\%)} & \cellcolor{gray!10}\textbf{0\% (4\%)} & \cellcolor{gray!10}\textbf{0\% (0\%)} & \cellcolor{gray!10}\textbf{0\% (0\%)}  \\
    \bottomrule
    \end{tabular}
    }
    \caption{ASR-GPT (outer) and ASR-keyword (in parentheses) for different defense methods on AdvBench. The best results are in \textbf{bold}. SafeInt outperforms all baselines across various attacks. 
    }
    \label{tab:defense_advbench}
\end{table*}

\begin{table*}[t]
    \centering
    \resizebox{\textwidth}{!}{
    \begin{tabular}{c l | c | c c c c c c}
    \toprule
        \multirow{2}{*}{Model} & \multirow{2}{*}{Defense}  & \multicolumn{1}{c|}{Benchmark $\downarrow$} & \multicolumn{6}{c}{Jailbreak Attacks $\downarrow$} \\
        & & JailbreakBench & GCG & AutoDAN & DeepInception & PAIR & TAP & MG \\
    \midrule
    
    \multirow{8}{*}{Vicuna}
    & No Defense        
        & 6\% (10\%) & 74\% (96\%) & 76\% (98\%) & 54\% (100\%) & 42\% (46\%) & 66\% (68\%) & 30\% (76\%)  \\
    & PPL               
        & 6\% (10\%) & 20\% (30\%) & 48\% (62\%) & 46\% (100\%) & 38\% (48\%) & 66\% (68\%) & 26\% (66\%) \\
    & Paraphrase
        & 6\% (20\%) & 18\% (40\%) & 22\% (60\%) & 28\% (98\%) & 16\% (36\%) & 32\% (42\%) & 10\% (40\%)  \\
    & Self-Examination        
        & 0\% (4\%) & 8\% (28\%) & 26\% (48\%) & 28\% (74\%) & 10\% (14\%) & 28\% (30\%) & 8\% (50\%) \\
    & ICD      
        & 0\% (0\%) & 8\% (14\%) & 42\% (42\%) & 54\% (94\%) & 16\% (28\%) & 32\% (42\%) & 22\% (52\%) \\
    & Self-Reminder
        & 0\% (2\%) & 4\% (4\%) & 4\% (6\%) & 36\% (100\%) & 14\% (20\%) & 26\% (30\%) & 30\% (62\%) \\
    & SafeDecoding              
        & 0\% (0\%) & 0\% (0\%) & 18\% (18\%) & 0\% (0\%) & 10\% (14\%) & 10\% (12\%) & 14\% (36\%) \\
    & \cellcolor{gray!10}\textbf{SafeInt (Ours)}     
        & \cellcolor{gray!10}\textbf{0\% (0\%)} & \cellcolor{gray!10}\textbf{0\% (0\%)} & \cellcolor{gray!10}\textbf{4\% (6\%)} & \cellcolor{gray!10}\textbf{0\% (0\%)} & \cellcolor{gray!10}\textbf{2\% (12\%)} & \cellcolor{gray!10}\textbf{4\% (12\%)} & \cellcolor{gray!10}\textbf{8\% (24\%)}  \\
    \midrule
    
    \multirow{8}{*}{Llama2}
    & No Defense        
        & 0\% (0\%) & 24\% (34\%) & 30\% (42\%) & 2\% (2\%) & 0\% (6\%) & 6\% (6\%) & 0\% (4\%) \\
    & PPL               
        & 0\% (0\%) & 2\% (2\%) & 6\% (6\%) & 2\% (2\%) & 0\% (4\%) & 6\% (6\%) & 0\% (2\%)  \\
    & Paraphrase
        & 0\% (6\%) & 0\% (28\%) & 0\% (22\%) & 2\% (2\%) & 0\% (24\%) & 4\% (28\%) & 0\% (10\%) \\
    & Self-Examination        
        & 0\% (0\%) & 2\% (2\%) & 6\% (6\%) & 0\% (0\%) & 0\% (4\%) & 2\% (2\%) & 0\% (2\%) \\
    & ICD      
        & 0\% (0\%) & 0\% (0\%) & 0\% (0\%) & 0\% (0\%) & 0\% (0\%) & 0\% (0\%) & 0\% (0\%)  \\
    & Self-Reminder
        & 0\% (0\%) & 0\% (0\%) & 0\% (0\%) & 0\% (0\%) & 0\% (0\%) & 0\% (0\%) & 0\% (10\%)  \\
    & SafeDecoding              
        & 0\% (0\%) & 0\% (0\%) & 0\% (0\%) & 0\% (0\%) & 0\% (2\%) & 0\% (6\%) & 0\% (0\%) \\
    & \cellcolor{gray!10}\textbf{SafeInt (Ours)}     
        & \cellcolor{gray!10}\textbf{0\% (0\%)} & \cellcolor{gray!10}\textbf{0\% (0\%)} & \cellcolor{gray!10}\textbf{0\% (0\%)} & \cellcolor{gray!10}\textbf{0\% (0\%)} & \cellcolor{gray!10}\textbf{0\% (6\%)} & \cellcolor{gray!10}\textbf{0\% (0\%)} & \cellcolor{gray!10}\textbf{0\% (0\%)}  \\
    \bottomrule
    \end{tabular}
    }
    \caption{ASR-GPT (outer) and ASR-keyword (in parentheses) on JailbreakBench. The best results are in \textbf{bold}. SafeInt consistently achieves the best performance. 
    }
    \label{tab:defense_jbb}
\end{table*}

\begin{table*}[t]
    \centering
    \resizebox{0.9\textwidth}{!}{
    \begin{tabular}{c l | c | c c c c c c}
        \toprule
            \multirow{2}{*}{Model} & \multirow{2}{*}{Defense} & \multirow{2}{*}{MT-Bench $\uparrow$} 
            & \multicolumn{6}{c}{Just-Eval $\uparrow$} \\  
            & & & Helpfulness & Clear & Factual & Deep & Engaging & Average \\
        \midrule
            \multirow{5}{*}{Vicuna} 
            & No Defense        
                & 5.21 & 4.44 & 4.66 & 4.38 & 3.60 & 3.49 & 4.11 \\
            & Self-Examination  
                & \underline{5.03} & 4.40 & 4.65 & 4.34 & 3.56 & 3.47 & \textbf{4.08} \\
            & ICD               
                & 4.86 & 4.34 & 4.61 & 4.34 & 3.40 & 3.32 & 4.00 \\
            & SafeDecoding      
                & 4.84 & 3.92 & 4.45 & 4.19 & 3.24 & 3.25 & 3.81 \\
            & \cellcolor{gray!10}SafeInt (Ours)            
                & \cellcolor{gray!10}\textbf{5.09} & \cellcolor{gray!10}4.40 & \cellcolor{gray!10}4.64 & \cellcolor{gray!10}4.35 & \cellcolor{gray!10}3.49 & \cellcolor{gray!10}3.41 & \cellcolor{gray!10}\underline{4.06} \\
        \midrule
            \multirow{5}{*}{Llama2} 
            & No Defense        
                & 5.80 & 4.65 & 4.78 & 4.50 & 4.19 & 3.90 & 4.40 \\
            & Self-Examination  
                & 1.61 & 3.21 & 3.67 & 3.47 & 2.92 & 2.68 & 3.19 \\
            & ICD               
                & 2.91 & 3.44 & 4.08 & 3.96 & 3.25 & 3.24 & 3.59 \\
            & SafeDecoding      
                & \underline{5.68} & 4.53 & 4.73 & 4.42 & 4.05 & 3.83 & \underline{4.31} \\
            & \cellcolor{gray!10}SafeInt (Ours)          
                & \cellcolor{gray!10}\textbf{5.82} & \cellcolor{gray!10}4.62 & \cellcolor{gray!10}4.76 & \cellcolor{gray!10}4.47 & \cellcolor{gray!10}4.13 & \cellcolor{gray!10}3.89 & \cellcolor{gray!10}\textbf{4.37} \\
        \bottomrule
    \end{tabular}
    }
    \caption{Utility evaluation scores of SafeInt and baselines. The highest and second-highest scores obtained by defense methods are in \textbf{bold} and \underline{underlined}, respectively. SafeInt maintains the best utility in most cases.
    }
    \label{tab:utility_eval}
\end{table*}

\subsection{Main Results}
\autoref{tab:defense_advbench} presents the ASR results of SafeInt and various baselines on AdvBench. For both Vicuna and Llama2, SafeInt achieves the best performance, reducing ASR to the lowest level among all defense methods under different attacks. Although our training process only utilizes jailbreak samples constructed with GCG, SafeInt effectively defends against other attack strategies, such as PAIR and TAP, which generate adversarial prompts using the LLM. This highlights the generalization capability of our defense, validating our previous observation. Moreover, even against MG attacks, SafeInt significantly lowers ASR, showing that it can generalize to different languages. 

\autoref{tab:defense_jbb} reports results on another out-of-distribution test set, JailbreakBench. SafeInt continues to outperform all baselines across different models and attack strategies. This demonstrates its robustness to unseen data.

While delivering strong defense performance, SafeInt largely preserves the utility of LLMs. As shown in \autoref{tab:utility_eval}, SafeInt achieves almost identical scores to the non-defended model in Llama2, whereas ICD and Self-Examination severely degrade utility. For Vicuna, SafeInt results in only a 2\% decrease in MT-Bench and a 1\% decrease in Just-Eval compared to the non-defended model. In contrast, SafeDecoding leads to 7\% drops in both benchmarks. 

Since our intervention essentially involves an incremental computation, it can be integrated directly into the forward propagation of the model. Unlike SafeDecoding, which requires an additional expert model for contrastive decoding, SafeInt introduces virtually no extra computational overhead and is therefore more efficient.

\begin{table}[t]
    \centering
    \resizebox{\columnwidth}{!}{
    \begin{tabular}{lcc}
        \toprule
        Jailbreak Attacks & AdvBench & JailbreakBench \\
        \midrule
        Adaptive-GCG & 0\% (0\%) & 0\% (6\%) \\
        Adaptive-AutoDAN & 0\% (0\%) & 6\% (8\%) \\
        \bottomrule
    \end{tabular}
    }
    \caption{Experimental results of defending against adaptive attacks on Vicuna, with evaluation metrics ASR-GPT and ASR-keyword (in parentheses). 'Adaptive-GCG' and 'Adaptive-AutoDAN' refer to GCG and AutoDAN attacks that are optimized in real-time based on the LLM deployed with SafeInt.}
    \label{tab:defense_adaptive}
\end{table}

\subsection{Adaptive Attack}
We also consider a scenario where the attacker knows SafeInt and has access to the LLM deployed with it. This means the attacker can dynamically adjust their attack strategies based on the latest defended LLM. The experimental results in \autoref{tab:defense_adaptive} show that SafeInt maintains strong defense performance under such adaptive settings.
Because SafeInt acts inside a low-rank subspace that exists only within the model, no explicit clues about this subspace are exposed at the prompt level or in the externally observable gradients. To bypass SafeInt, an attacker must precisely manipulate specific components of the original representations within this subspace. Therefore, even if GCG and AutoDAN optimize their adversarial prompts in real time, it is difficult to generate threatening attacks.

\section{Analyses}

\subsection{Ablation Studies}

We conduct ablation studies on the introduced contrastive loss and reconstruction loss to verify their effectiveness. As shown in \autoref{tab:ablation}, removing contrastive loss increases ASR, indicating its crucial role in enhancing defense performance. Incorporating contrastive loss leads to a decrease in MT-Bench scores, which may be attributed to its impact on the overall representation structure when pulling together or pushing apart local representations. When the reconstruction loss is omitted, representations are more susceptible to excessive intervention, resulting in both defense failure and a significant decline in response quality.

\begin{table}[t]
    \centering
    \resizebox{\columnwidth}{!}{
    \begin{tabular}{lcccc}
        \toprule
        \multirow{2}{*}{Methods} & \multicolumn{3}{c}{Jailbreak Attacks $\downarrow$} & \multirow{2}{*}{MT-Bench $\uparrow$} \\
         & GCG & AutoDAN & PAIR &  \\
        \midrule
        No Defense & 90\% & 82\% & 54\% & 5.21 \\
        SafeInt & 0\% & 2\% & 2\% & 5.09 \\
        \midrule
        w/o $\mathcal{L}_{ct}$ & 2\% & 8\% & 6\% & 5.22 \\
        w/o $\mathcal{L}_{recon}$ & 2\% & 12\% & 8\% & 4.09 \\
        \bottomrule
    \end{tabular}
    }
    \caption{Ablation results of our method on AdvBench and MT-Bench, using ASR-GPT as the metric for Jailbreak Attacks. 'w/o $\mathcal{L}_{ct}$' and 'w/o $\mathcal{L}_{recon}$' denote the removal of contrastive loss and reconstruction loss, respectively.
    }
    \label{tab:ablation}
\end{table}

\begin{figure}[t]
    \centering
    \begin{subfigure}{0.8\columnwidth}
        \centering
        \includegraphics[width=\textwidth]{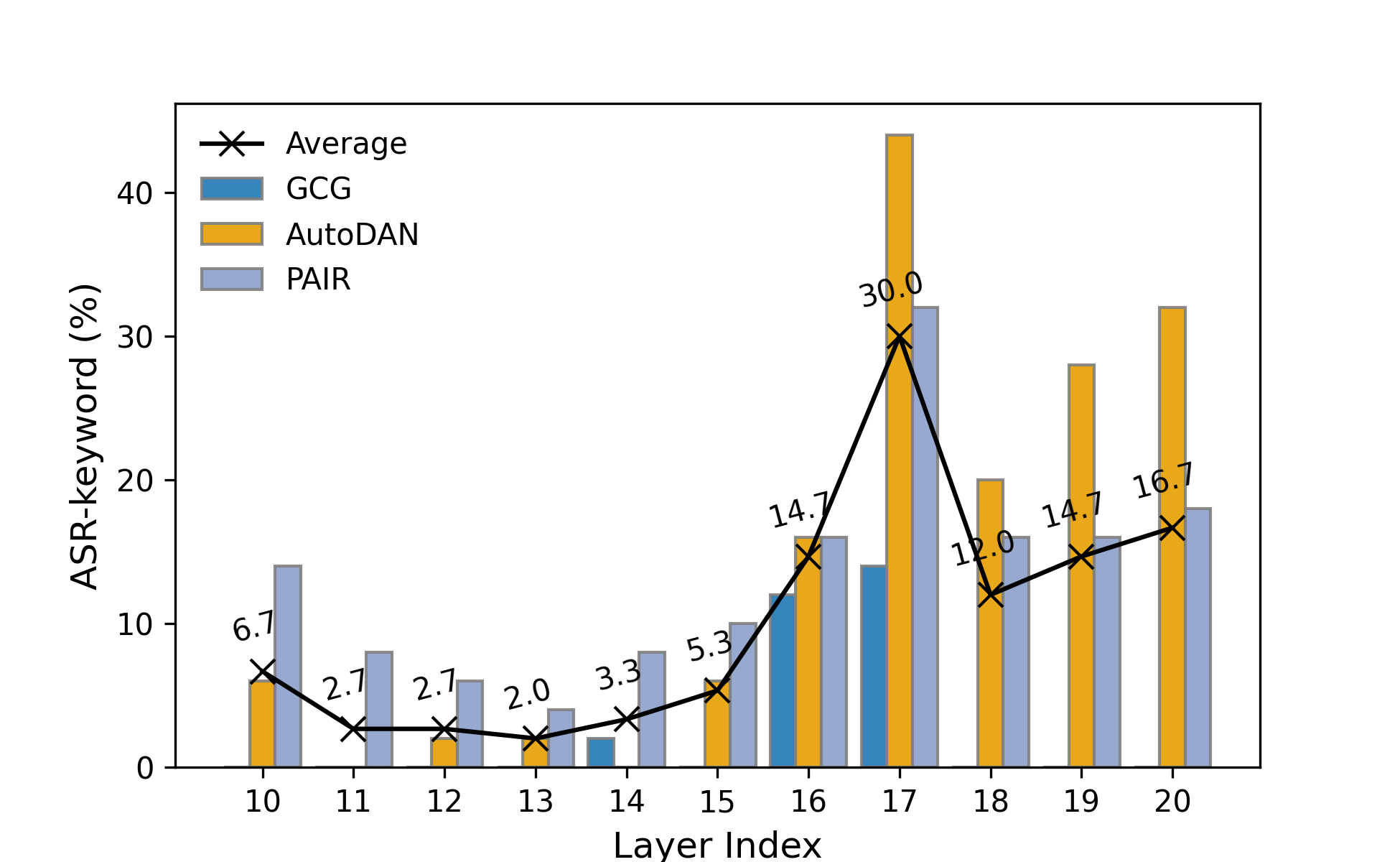}
        \caption{Intervention Layer $\mathcal{I}$}
        \label{fig:intervention_layer}
    \end{subfigure}
    \hfill
    \begin{subfigure}{0.8\columnwidth}
        \centering
        \includegraphics[width=\textwidth]{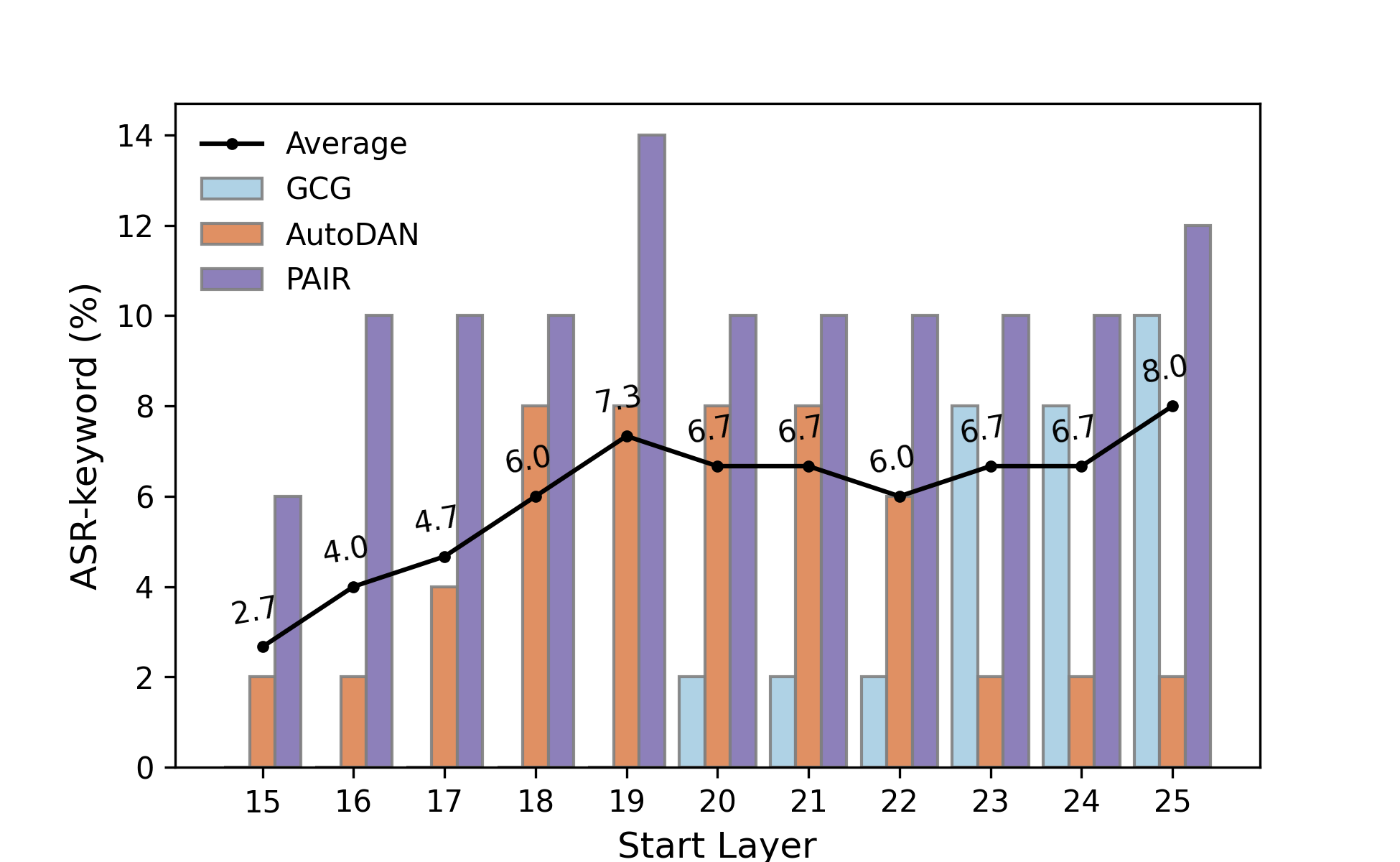}
        \caption{Alignment Layer Range $\mathbb{L}^{a}$}
        \label{fig:alignment_range}
    \end{subfigure}
    \caption{Analysis of the intervention layer and alignment layer range.}
    \label{fig:hypa_analysis}
\end{figure}

\subsection{Hyperparameter Analysis}    
\textbf{Intervention Layer Choice}\quad 
To understand how the choice of the intervention layer impacts our defense effectiveness, we conduct an analysis. \autoref{fig:hypa_analysis}(\subref{fig:intervention_layer}) displays the ASR-keyword when the intervention layer is set between layers 10 and 20. We observe that intervening in the intermediate layers generally yields better results than intervening in the later layers, which may suggest that these intermediate layers play a more crucial role in jailbreak mechanisms. Notably, when the intervention is applied at layer 13, ASR reaches its lowest point. This finding aligns with our observation in \autoref{fig:cls_ood}, where Vicuna exhibits the highest jailbreak representation consistency at layer 13.
\vspace{0.2\baselineskip}
\\ \textbf{Alignment Layer Range}\quad 
We fix the endpoint of the alignment layer range at the final layer and modify the starting point to control its span. In \autoref{fig:hypa_analysis}(\subref{fig:alignment_range}), we illustrate the results when setting the starting point between layers 15 and 25. We observe that as the starting point shifts to later layers, the defense effectiveness weakens. This may be attributed to the reduced number of aligned layers being insufficient to correct the attack. Overall, while adjusting the alignment layer range impacts defense performance, the effect is not drastic, indicating that our method exhibits a certain degree of robustness to this hyperparameter.


\section{Related Work}
\textbf{Jailbreak Attacks}\quad
Jailbreak attacks aim to bypass alignment or safeguards, forcing LLMs to generate inappropriate content. Early jailbreak attacks \cite{wei2023jailbroken, yong2024lowresource, yuan2024cipher} rely on manually crafted adversarial prompts, which primarily exploit objective competition and mismatched generalization to achieve jailbreaks. Subsequent optimization-based attacks \cite{zou2023universal, liu2024autodan, paulus2024advprompter} introduce automated adversarial prompt optimization by leveraging the internal states of LLMs, significantly improving both the success rate and efficiency of jailbreaks. Recent jailbreak attacks \cite{chao2024jailbreaking, mehrotra2024tree, ding2024wolf} iteratively rewrite and refine adversarial prompts using one or multiple LLMs, further exposing security vulnerabilities in LLMs.
\vspace{0.2\baselineskip}
\\ \textbf{Jailbreak Defenses}\quad 
To address the challenges posed by jailbreak attacks, numerous defense methods have been proposed \cite{robey2024smoothllm, kumar2025certifying}. Detection-based approaches identify adversarial prompts by computing perplexity \cite{alon2023detecting} or randomly deleting parts of the input \cite{cao2024rallm}. Some methods prompt the LLM to perform self-checking through instructions \cite{phute2024selfexam, xie2023defending, zhang2024gp} or context \cite{zhou2024icag}. Decoding-based defenses \cite{xu-etal-2024-safedecoding, liu2024aed} focus on analyzing decoding probabilities under different conditions and formulating decoding strategies to ensure safer outputs. Additionally, certain approaches \cite{zhao2024LED, ouyang2025layer} edit specific model parameters to make LLMs forget harmful knowledge. A more controllable and efficient class of defenses \cite{li2025revisiting, shen2025JA} involves manipulating representations to mitigate jailbreak attacks without modifying model parameters or adding decoding overhead. 
\vspace{0.2\baselineskip}
\\ \textbf{Representation Engineering for Safety}\quad     
Many studies have employed representation engineering techniques \cite{zou2023RepE} to investigate or enhance the safety of LLMs. \citet{zhou2024explain} and \citet{arditi2024refusal} analyze the mechanisms of jailbreak and refusal from a representation perspective, respectively. 
\citet{li2025revisiting} improve the robustness of LLMs by strengthening the safety patterns they recognize.
\citet{zheng2024DRO} introduce a learnable safety prompt that aims to increase the separation between harmful and harmless query representations along the refusal direction. 
\citet{shen2025JA} add a difference vector to query representations to guide the LLM toward rejecting malicious instructions, while \citet{gao2024shaping} mitigate jailbreak attacks by constraining activations within a safe boundary. 
A major drawback of these two approaches is that their interventions cannot be automatically optimized. This means that when the intervention is applied to all query representations, the choice of intervention strength becomes highly sensitive.   
In contrast, our method adopts a parameterized intervention, which adaptively identifies and adjusts jailbreak-related representations regardless of manually tuning the intervention strength.

\section{Conclusion}
This paper first analyzes the representations of jailbreak samples on four LLMs and makes key observations. Building on these observations, we propose SafeIntervention (SafeInt), a novel method that defends LLMs against jailbreak attacks via representation intervention. SafeInt can adaptively identify and intervene in jailbreak-related representations while seamlessly integrating into the LLM inference process.
Comprehensive experimental results show that our proposed SafeInt outperforms all baselines in defending against jailbreak attacks. In most cases, SafeInt also achieves the best utility.


\section*{Limitations}
We discuss the limitations of our work. We make a preliminary observation that SafeInt can transfer to different but homologous LLMs. We speculate that these homologous LLMs may share similar jailbreak representation structures. However, we have not conducted an in-depth exploration of the transferability of SafeInt.

\bibliography{custom}

\newpage
\appendix

\section{Appendix} \label{sec:appendix}
\subsection{Dataset Details}   \label{sec:dataset_details}
\vspace{0.2\baselineskip}
\subsubsection{Training Data}
\vspace{0.3\baselineskip}
To construct $ \mathcal{D}^{(\text{train})}_{\text{jailbreak}} $, we use GCG to generate jailbreak instructions from 128 randomly sampled instructions from AdvBench \cite{zou2023universal}.  
\vspace{0.3\baselineskip}
\\ To construct $ \mathcal{D}^{(\text{train})}_{\text{unsafe}} $, we sample 128 harmful instructions from MaliciousInstruct \cite{huang2023catastrophic} and TDC2023 \cite{tdc2023}.  
\vspace{0.3\baselineskip}
\\ To construct $ \mathcal{D}^{(\text{train})}_{\text{safe}} $, we sample 128 harmless instructions from Alpaca \cite{alpaca}.

\subsubsection{Test Data in \textit{Q1}}
\vspace{0.3\baselineskip}
To construct $ \mathcal{D}^{(\text{test})}_{\text{jailbreak}} $, we first resample 150 instructions from AdvBench. We then use GCG to generate jailbreak instructions from these 150 instructions.  
\vspace{0.3\baselineskip}
\\ To construct $ \mathcal{D}^{(\text{test})}_{\text{unsafe}} $, we sample 150 harmful instructions from Do Not Answer \cite{wang2023donotanswer}, MaliciousInstruct, and TDC2023.\footnote{Due to the insufficient data volume of the original two datasets, we introduce Do Not Answer.}
\vspace{0.3\baselineskip}
\\ To construct $ \mathcal{D}^{(\text{test})}_{\text{safe}} $, we sample 150 harmless instructions from Alpaca.  
\vspace{0.4\baselineskip}
\\ Note that $ \mathcal{D}^{(\text{test})}_{\text{jailbreak}} $, $ \mathcal{D}^{(\text{test})}_{\text{unsafe}} $, and $ \mathcal{D}^{(\text{test})}_{\text{safe}} $ do not overlap with $ \mathcal{D}^{(\text{train})}_{\text{jailbreak}} $, $ \mathcal{D}^{(\text{train})}_{\text{unsafe}} $, and $ \mathcal{D}^{(\text{train})}_{\text{safe}} $.

\vspace{0.2\baselineskip}
\subsubsection{Test Data in \textit{Q2}}
\vspace{0.2\baselineskip}
To construct $ \mathcal{D}^{({\scriptscriptstyle \text{OOD}})}_{\text{jailbreak}} $, we first sample 50 instructions from AdvBench. We then use three jailbreak methods—GCG, AutoDAN, and DeepInception—to generate jailbreak instructions from these 50 samples, resulting in a total of 150 jailbreak instructions.  
\vspace{0.3\baselineskip}
\\ For unsafe and safe samples, we continue to use $ \mathcal{D}^{(\text{test})}_{\text{unsafe}} $ and $ \mathcal{D}^{(\text{test})}_{\text{safe}} $.  
\vspace{0.4\baselineskip}
\\ Note that $ \mathcal{D}^{({\scriptscriptstyle \text{OOD}})}_{\text{jailbreak}} $ does not overlap with $ \mathcal{D}^{(\text{test})}_{\text{jailbreak}} $ or $ \mathcal{D}^{(\text{train})}_{\text{jailbreak}} $.

\vspace{0.2\baselineskip}
\subsection{Attack Setup}   \label{sec:attack_setup}
For GCG, AutoDAN, DeepInception, PAIR, and TAP, we follow the settings from their original papers. For multilingual mismatch generalization attacks, we translate each instruction in the test set into one of six non-English languages—Chinese, Italian, Vietnamese, Arabic, Korean, or Thai—to carry out the attacks.

\subsection{Details of ASR-keyword}
\autoref{tab:refusal_strings} lists the refusal keywords used to evaluate ASR-keyword. We adopt the settings from \citet{xu-etal-2024-safedecoding} and further refine them by adding "I am not programmed to" and "As a responsible". Additionally, we exclude "I'm an" and "I'm just" to prevent false positives.

\begin{table}[h]
    \centering
    \begin{tabular}{c}
        \toprule
        I'm sorry \\
        I am sorry \\
        I apologize \\
        As an AI \\
        As a responsible \\
        As a language model \\
        As an assistant \\
        I cannot \\
        I can't \\
        I am unable to \\
        I'm unable to \\
        I'm just an AI \\
        I do not \\
        It is not</s> \\
        It is not appropriate \\
        I am not allowed to \\
        I am not programmed to \\
        I am an AI language model \\
        \bottomrule
    \end{tabular}
    \caption{Refusal Keywords}
    \label{tab:refusal_strings}
\end{table}








\end{document}